\documentclass[letterpaper, 10 pt, conference]{ieeeconf}  

\IEEEoverridecommandlockouts                              

\usepackage[cmex10]{amsmath}
\usepackage{amsfonts}
\usepackage{amsthm}

\usepackage{multicol}

\usepackage{xspace}

\newcommand\moduleName[1]{\textsf{#1}\xspace}
\newcommand{\MAMO}{\moduleName{MAMO}}
\newcommand{\MAPF}{\moduleName{MAPF}}
\newcommand{\MfM}{\moduleName{\fontfamily{pbk}\selectfont M{\footnotesize 4}M}}

\newcommand{\CBS}{\textsc{CBS}\xspace}

\newcommand{\KPIECE}{\textsc{KPIECE}\xspace}
\newcommand{\RRT}{\textsc{RRT}\xspace}

\newcommand{\SelSim}{\textsc{SelSim}\xspace}
\newcommand{\Dogar}{\textsc{Dogar}\xspace}

\usepackage[table,xcdraw,dvipsnames,x11names]{xcolor}
\usepackage{color}

\colorlet{documentLinkColor}{red}
\colorlet{documentUrlColor}{blue}
\colorlet{documentCitationColor}{ForestGreen}
\usepackage[bookmarks=true]{hyperref}
\hypersetup{
     colorlinks = true,
     citecolor = documentCitationColor,
     linkcolor = documentLinkColor,
     urlcolor = documentUrlColor,
}

\usepackage{graphicx}

\newcommand{\PP}{\mathcal{P}}
\newcommand{\X}{\mathcal{X}}
\newcommand{\A}{\mathcal{A}}
\newcommand{\T}{\mathcal{T}}

\newcommand{\R}{\mathcal{R}}
\newcommand{\Obs}{\mathcal{O}}
\newcommand{\OI}{\mathcal{O}_I}
\newcommand{\OM}{\mathcal{O}_M}

\DeclareMathOperator*{\argmin}{\arg\!\min}

\usepackage{textcomp,gensymb}

\usepackage{algorithm}
\usepackage[noend]{algpseudocode} 
\let\oldReturn\Return
\renewcommand{\Return}{\State\oldReturn}

\theoremstyle{proposition}

\usepackage{titlesec}

\usepackage[binary-units]{siunitx}
\usepackage{booktabs}
\usepackage{multirow}

\usepackage{geometry}
\usepackage{cite}

\usepackage{mathtools}
\usepackage{balance}

\title{\LARGE \bf
Planning for Complex Non-prehensile Manipulation Among Movable Objects by Interleaving Multi-Agent Pathfinding and Physics-Based Simulation
}

\author{Dhruv Mauria Saxena$^{1}$ and Maxim Likhachev$^{1}$
\thanks{$^{1}$The authors are with the
Robotics Institute, Carnegie Mellon University, Pittsburgh, PA 15213, USA.
{\small e-mail: \tt \{dsaxena, mlikhach\}@andrew.cmu.edu}. This work was in part supported by ARL grant W911NF-18-2-0218 and ONR grant N00014-18-1-2775.}%
}

\begin{document}

\maketitle
\thispagestyle{empty}
\pagestyle{empty}


\begin{abstract}

Real-world manipulation problems in heavy clutter require robots to reason about potential contacts with objects in the environment.
We focus on pick-and-place style tasks to retrieve a target object from a shelf where some `movable' objects must be rearranged in order to solve the task.
In particular, our motivation is to allow the robot to reason over and consider non-prehensile rearrangement actions that lead to complex robot-object and object-object interactions where multiple objects might be moved by the robot simultaneously, and objects might tilt, lean on each other, or topple.
To support this, we query a physics-based simulator to forward simulate these interaction dynamics which makes action evaluation during planning computationally very expensive.
To make the planner tractable, we establish a connection between the domain of \textbf{\fontfamily{pbk}\selectfont M}anipulation Among Movable Objects and \textbf{\fontfamily{pbk}\selectfont M}ulti-Agent Pathfinding that lets us decompose the problem into two phases our \MfM algorithm iterates over.
First we solve a multi-agent planning problem that reasons about the configurations of movable objects but does not forward simulate a physics model.
Next, an arm motion planning problem is solved that uses a physics-based simulator but does not search over possible configurations of movable objects.
We run simulated and real-world experiments with the PR2 robot and compare against relevant baseline algorithms.
Our results highlight that \MfM generates complex 3D interactions, and solves at least twice as many problems as the baselines with competitive performance.

\end{abstract}


\section{Introduction}

Manipulation Among Movable Objects \raisebox{0.5pt}{(}\MAMO{\raisebox{0.5pt}{)}}~\cite{StilmanMAMO} defines a broad class of problems where a robot must complete a manipulation task in the presence of obstructing clutter.
In heavily cluttered scenes, there may be no collision-free trajectory that solves the task.
This does not make the problem unsolvable since \MAMO allows rearrangement of some objects \emph{a priori} designated as `movable'.
In addition, \MAMO may associate each object with constraints on how it can be interacted with -- it is undesirable to allow robots to carelessly push or throw objects around.

\begin{figure}[t]
    \centering
    \includegraphics[width=1.0\columnwidth]{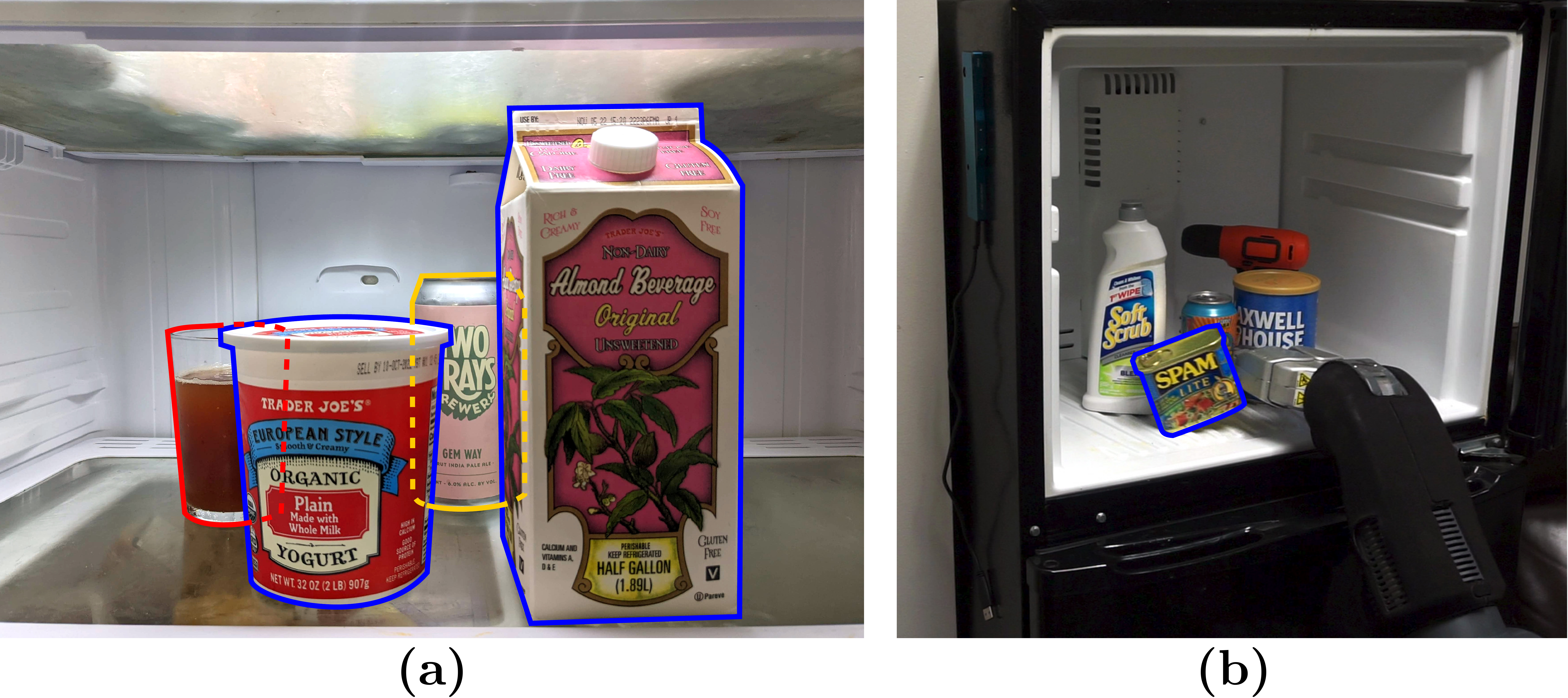}
    \caption{(a) An example \MAMO problem to retrieve the beer can (yellow outline). Access is blocked by the movable box of milk and tub of yogurt (blue outlines). In order to retrieve the can, they must be rearranged out of the way without toppling them, and without anything making contact with the glass of juice (red outline). (b) A complex non-prehensile action that tilts the movable potted meat can (blue outline) to rearrange it.}
    \label{fig:intro_fridge}
\end{figure}

In this paper, we consider \MAMO problems for pick-and-place manipulation tasks where the robot needs to retrieve a target object from a cluttered shelf, cabinet, fridge, or a similar structure.
Fig.~\ref{fig:intro_fridge} (a) shows an example of such a scene where two movable objects must be rearranged in order to retrieve the desired object, while ensuring they do not topple and no contacts are made with an immovable obstacle.

Solving such \MAMO problems requires answers to three difficult questions: \emph{which} objects to move, \emph{where} to move them, and \emph{how} to move them.
Thus \MAMO problems assign the robot a goal with respect to the overall task and object-of-interest (OoI), without any additional goal specifications for other objects except for satisfying their associated interaction constraints; while \MAMO solutions exist in a composite configuration space that includes the configuration of the robot arm and all objects in the scene.
The search for a solution is computationally challenging since the size of this space grows exponentially with the number of objects.


We are interested in non-prehensile rearrangement actions since they allow robots to manipulate objects that may be too big or too bulky or otherwise ungraspable.
In many cases it is more time- and energy-efficient to push an object off to the side than to grasp it, pick it up, move it elsewhere, place it down, and release it before proceeding.
Furthermore, we allow the robot to move multiple objects simultaneously with the same push action, and we allow objects to tilt, lean on each other, and slide (an example is shown in Fig.~\ref{fig:intro_fridge} (b)).
Planning with these actions requires the ability to predict the effect of robot actions on the configuration of objects, typically through computationally expensive forward simulations of a rigid-body physics simulator.

\begin{figure}[t]
    \centering
    \includegraphics[width=1.0\columnwidth]{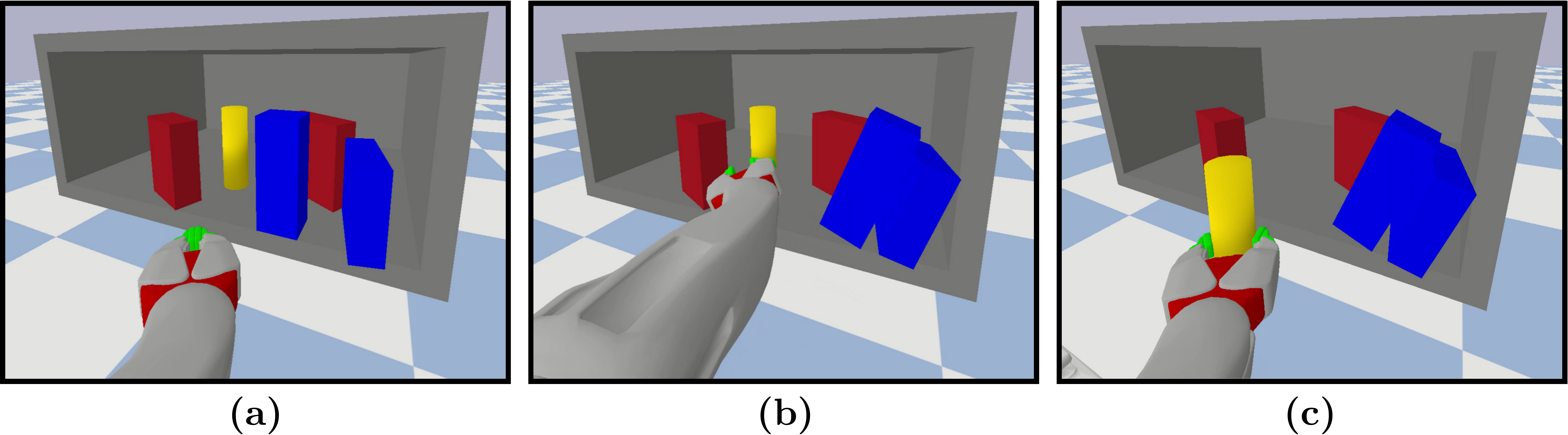}
    \caption{Sequence of images showing a solution found by our \MfM algorithm for a simple \MAMO scene. From \emph{left} to \emph{right}: (a) initial scene, (b) rearranged scene after one push action, (c) successful OoI retrieval. Movable objects are blue, immovable obstacles are red, and the OoI is yellow.}
    \label{fig:intro_soln}
\end{figure}

Our key insight in this work draws a connection between the \MAMO domain and Multi-Agent Pathfinding \raisebox{0.5pt}{(}\MAPF{\raisebox{0.5pt}{)}} to decompose the problem into two parts.
First, we treat the movable objects as artificially actuated agents tasked with avoiding collisions with (i) our robot arm retrieving the OoI, (ii) each other, and (iii) immovable obstacles.
A solution to this abstract \MAPF problem searches over potential rearrangements of objects \emph{without} the need to query a physics simulator.
Next, we use the \MAPF solution to compute informed push actions to rearrange movable objects \emph{without} searching over their possible configurations.
These actions are forward simulated with a physics model to ensure validity.
The decomposition helps us keep track of object configurations in the full $SE(3)$ space and generate informed push actions that lead to realistic multi-body interactions in the 3D workspace as shown in Fig.~\ref{fig:intro_fridge} (b). 
Fig~\ref{fig:intro_soln} shows a complex and interesting solution found by our algorithm for one of the simpler scenarios in our test data.

The main contributions of our work in this paper for solving \MAMO planning problems are:
\begin{itemize}
    \item Enable reasoning over and usage of complex non-prehensile interactions that may push multiple objects in tandem and produce object-object interactions like leaning and toppling  (Fig.~\ref{fig:intro_fridge} (b)).
	\item \MAPF abstraction for computing suitable rearrangements for \MAMO planning problems, without using a simulation-based model.
	\item An efficient algorithm to solve \MAMO problems that iterates between calls to an \MAPF solver (to determine \emph{which} objects to move \emph{where}) and a push planner (to verify \emph{how} to move the objects).
	\item A thorough experimental evaluation of our approach in simulation and in the real-world on a PR2 robot.
\end{itemize}

We provide details of relevant works from \MAMO literature in Section~\ref{sec:related_work}.
Section~\ref{sec:prelims} formalises the \MAMO planning problem. 
Section~\ref{sec:m4m} presents our iterative planning algorithm \MfM, including the abstraction from \MAMO to \MAPF (Section~\ref{sec:key_idea}) and a non-prehensile push planner (Section~\ref{sec:push_actions}). 
We provide extensive quantitative evaluation against relevant \MAMO baselines in simulation in Section~\ref{sec:exps} along with real-world results of our algorithm on the PR2 robot.
Section~\ref{sec:discussion} discusses the benefits, limitations, and future extensions of this work. 


\section{Related Work}\label{sec:related_work}



\MAMO generalises Navigation Among Movable Obstacles \raisebox{0.5pt}{(}\textsf{NAMO}\raisebox{0.5pt}{)} where a mobile robot must navigate from start to goal in a reconfigurable environment~\cite{alami,Wilfong91,StilmanNAMO}.
It is also related to the rearrangement planning problem~\cite{Ben-ShaharR98,Ota09} which explicitly specifies desired goal configurations for movable objects.
Wilfong~\cite{Wilfong91} showed that rearrangement planning is PSPACE-hard, and \MAMO problems are NP-hard to solve.


Many existing \MAMO and \emph{rearrangement planning} solvers make use of prehensile actions~\cite{StilmanMAMO,KrontirisSDKB14,KrontirisB15,ShomeB21,WangGNYB21_uniform_rearrangement,LeeCNPK19}.
This simplifies planning since grasped objects behave as rigid bodies attached to the robot, but assumes access to known stable configurations of and grasp poses for objects~\cite{StilmanMAMO,KrontirisSDKB14,KrontirisB15,ShomeB21}.
In some cases a ``buffer'' location to place grasped objects is required~\cite{WangGNYB21_uniform_rearrangement,LeeCNPK19}.
In particular,~\cite{KrontirisSDKB14} and~\cite{ShomeB21} utilise the concept of ``pebble graphs''~\cite{KMS84,SoloveykColor} from \MAPF literature to find prehensile actions for rearrangement planning.
Their formulation restricts the motion of the movable objects (pebbles) on a precomputed roadmap of robot arm trajectories via prehensile actions.
This limits the possible configurations of objects they consider since motions are limited to poses from where they can be grasped and to those where they can be stably placed.
Since we utilise non-prehensile pushes for rearrangement and a physics-based simulator for action validation, our planner explores a richer space of robot-object and object-object interactions in the 3D workspace. 

Allowing \emph{non-prehensile interactions} with objects typically requires access to a simulation model to obtain the result of complex interaction dynamics~\cite{BergSKLM08,DogarS12,King1,HuangHYB22,homology_nonprehensile,SelSim}.
Of these approaches, only Selective Simulation~\cite{SelSim} considers realistic interactions in the 3D workspace and is one of our comparative baselines in Section~\ref{sec:exps}.
Others rely on planar robot-object interactions which fail to account for object dynamics in $SE(3)$ where they might tilt, lean, or topple.
In Section~\ref{sec:exps}, we adapt the \MAMO solver from~\cite{DogarS12} to use our push actions that lead to 3D robot-object interactions and require a physics simulator during planning.
Originally their work was limited to interacting with a single object at a time, and used an analytical motion model in $SE(2)$ to propagate the effect of the push on the planar configuration of the object being pushed (tilting and toppling was not considered in~\cite{BergSKLM08,DogarS12,King1,HuangHYB22,homology_nonprehensile}).

Querying \emph{physics-based simulators} for the result of an action is much more expensive than collision checking it.
\KPIECE~\cite{KPIECE} is a randomised algorithm for planning with a computationally expensive transition model (querying a physics-based simulator is an example of such a model).
\KPIECE and RRT~\cite{RRT} are two other baselines we compare against in Section~\ref{sec:exps}.
In our own prior work on \MAMO planning~\cite{SPAMP}, we find a collision-free trajectory to a region near the OoI grasp pose, and simulate goal-directed non-prehensile actions only within this region.
The assumption that such a collision-free trajectory exists is easily violated in the cluttered \MAMO workspaces we instantiate in our experiments (see Figs.~\ref{fig:intro_fridge} (a),~\ref{fig:levels}, and~\ref{fig:pr2_soln} for example).

\section{Problem Statement}\label{sec:prelims}


Let $\X_\R \subset \mathbb{R}^q$ denote the configuration space of a $q$ degrees-of-freedom robot manipulator $\R$.
Let $\Obs = \{O_1, \ldots, O_n\}$ be the set of objects in the scene, and $\X_{O_i} \equiv SE(3)$ be the configuration space of object $O_i$ that includes its 3D position and orientation.
The search space for a \MAMO planning problem is
$\X = \X_\R \times \X_{O_1} \times \cdots \times \X_{O_n}$.
We denote movable objects by $\OM$ and immovable obstacles by $\OI$ such that $\Obs = \OM \cup \OI$ and $\OM \cap \OI = \emptyset$.

Each object is associated with a set of interaction constraints.
For example, an `immovable' obstacle (an object that cannot be interacted with, such as a wall) will contain a constraint function which is satisfied so long as neither the robot nor any other object makes contact with it.
In our problems similar functions encode that movable objects cannot fall off the shelf, tilt too far (beyond $\SI{25}{\degree}$), or move with a high instantaneous velocity (above $\SI{1}{\meter\per\second}$).
A state $x \in \X$ is valid if all constraints for all objects are satisfied at that state. 
Let $\X_V$ be the space of valid states.

A \MAMO planning problem can be defined with the tuple $\PP = (\X, \A, \T, c, x_S, \X_G)$.
$\A$ is the action space of the robot, $\T: \X \times \A \rightarrow \X$ is a deterministic transition function, $c: \X \times \X \rightarrow \mathbb{R}_{\geq 0}$ is a state transition cost function, $x_S \in \X_V$ is the start state, and $\X_G \subset \X, \X_G \cap \X_V \neq \emptyset$ is the set of goal configurations.
The start state $x_s$ includes a ``home'' robot configuration in $\X_\R$ and the initial poses of all objects. 
We would like to find the least-cost valid path $\pi^*$ from start to goal i.e., a path made up of a sequence of valid states.
Formally, we can write this as:

\begin{equation}\label{eq:background_problem}
\begin{aligned}
    \text{find } &\pi^* = \argmin_{\pi = \{x_1, \ldots, x_T\}} \sum_{i=1}^{T-1} c(x_i, x_{i+1}) \\
    \text{s.t. } &x \in \X_V, \,\forall\, x \in \pi \quad\qquad\qquad\emph{(path of valid states)} \nonumber \\
    &x_1 = x_S, x_T \in \X_G \,\qquad\qquad\emph{(start, goal constraints)} \nonumber \\
    &x_{i+1} = \T(x_i, a_i), a_i \in \A, \!\!\!\!\!\quad\forall x_i, x_{i+1} \in \pi \nonumber \\
    & \qquad\qquad\qquad\qquad\qquad\qquad\emph{(transition dynamics)}
\end{aligned}
\end{equation}

\begin{figure}[t]
    \centering
    \includegraphics[width=0.8\columnwidth]{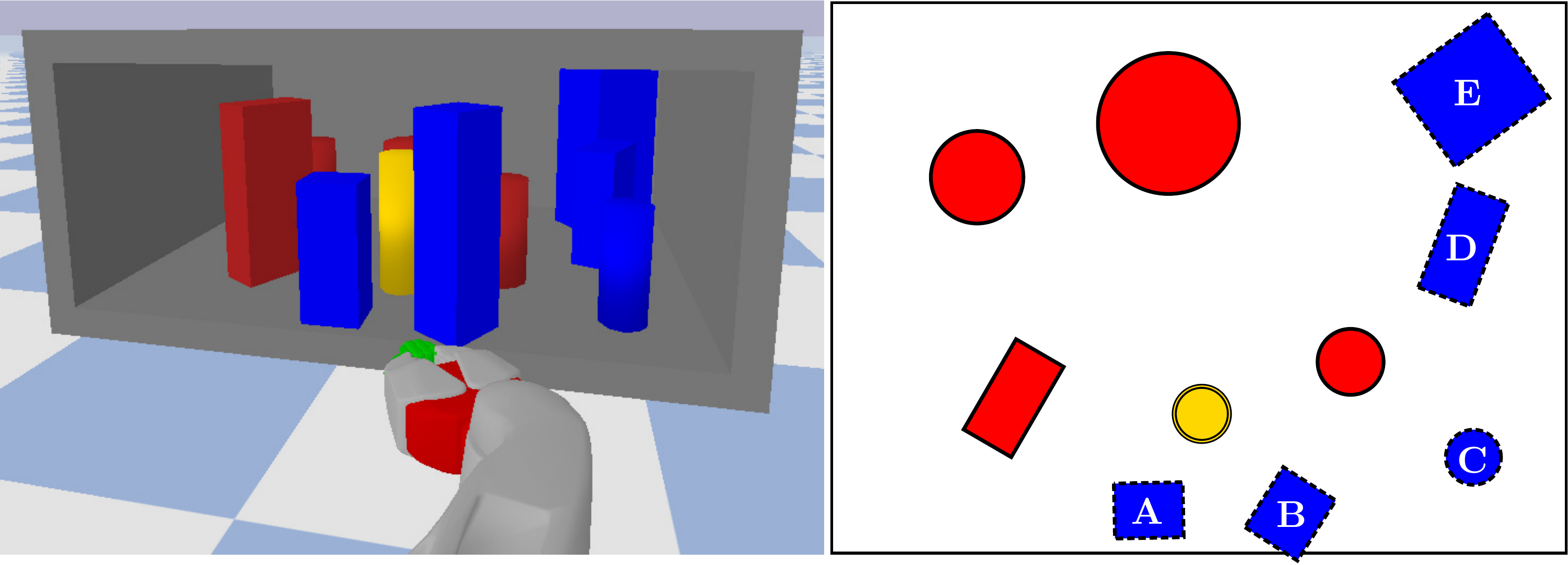}
    \caption{\MAMO workspace (\emph{left}) and its 2D projection labelled with movable object IDs. Movable objects are in blue, immovable obstacles in red, and the object-of-interest to be retrieved in yellow.}
    \label{fig:100011}
\end{figure}

In our work we discretise $\A$ to include ``simple motion primitives'' that independently change each robot joint angle by a fixed amount and dynamically generated ``push actions'' described in Section~\ref{sec:push_actions}. 
For transition $x_{i+1} = \T(x_i, a_i)$, action $a_i \in \A$ can affect object configurations between $x_i$ and $x_{i+1}$ only if $a_i$ is a push action or the OoI has been grasped.
The cost of robot actions $c$ is proportional to the distance travelled in $\X_\R$.
We assume $\X_G$ is defined in two parts -- a grasp pose in $SE(3)$ for the OoI and a goal pose in $SE(3)$ where it must end up (while grasped by the robot).
Our solution to \MAMO problems is a sequence of arm trajectories in the robot configuration space $\X_\R \subset \mathbb{R}^q$ ($q = 7$ for the PR2 robot) that (i) rearrange movable clutter and (ii) retrieve the OoI.
Fig.~\ref{fig:100011} shows an example of the \MAMO problems we consider in this paper, along with its 2D projection. Red objects are immovable obstacles $\OI$, blue objects are initial movable objects $\OM^{\text{init}}$, and the goal for the robot arm is to extract the yellow OoI from the shelf.
There is no collision-free trajectory for the arm to extract the OoI from the shelf.
Upon rearrangement of some movable objects ($A$ and $B$ in particular), such a trajectory may be found.

\section{The \MfM Planning Algorithm}\label{sec:m4m}

\begin{figure}[t]
    \centering
    \includegraphics[width=0.65\columnwidth]{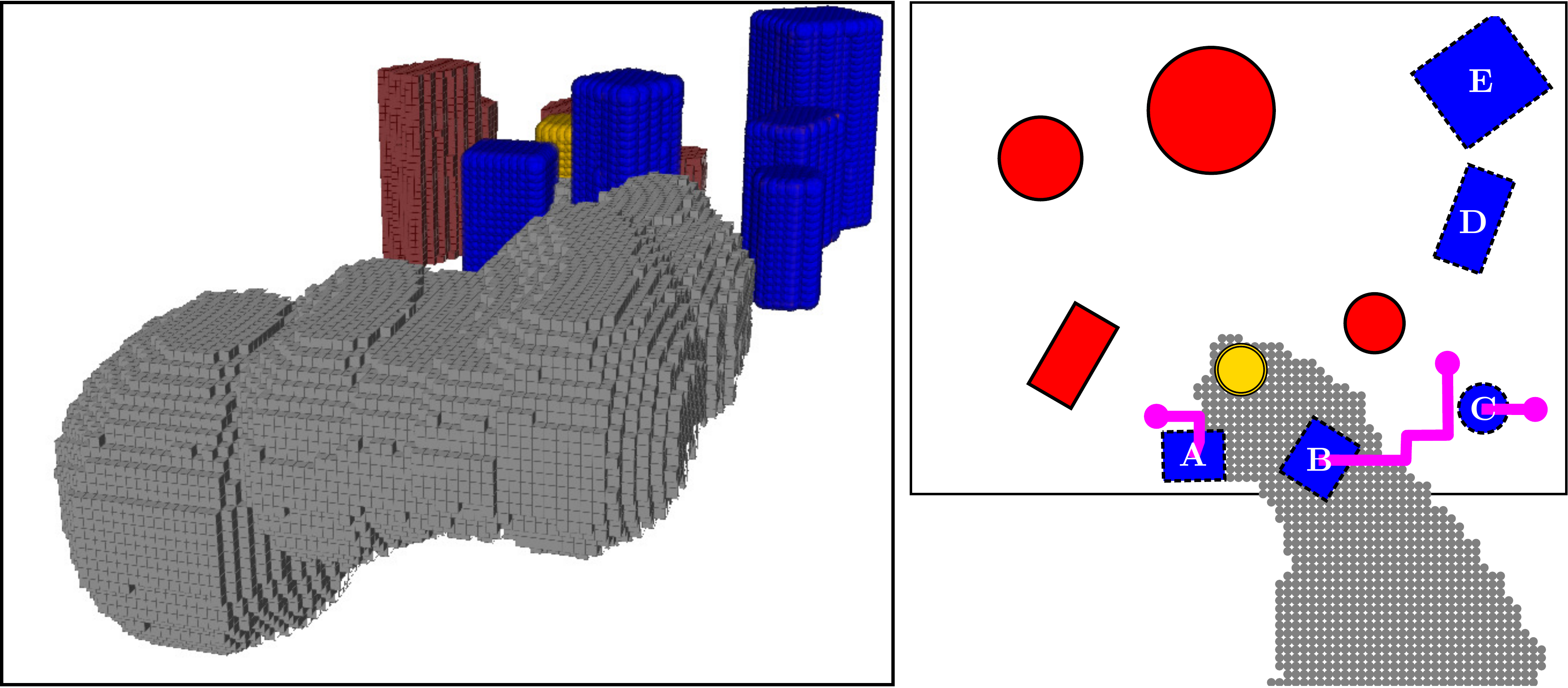}
    \caption{The negative goal region (NGR) $\mathcal{V}\left(\hat{\pi}_\R\right)$ in gray for the \MAMO problem from Fig.~\ref{fig:100011}. (\emph{left}) 3D volumes of the NGR and all objects at their initial poses (we omit the shelf for ease of visualisation). (\emph{right}) 2D projection of the NGR and the workspace, overlayed with the solution to the abstract \MAPF problem from Section~\ref{sec:key_idea} formulated for this scene. Objects $A$ and $B$ need to move outside the NGR, and object $C$ needs to move to allow $A$ to reach its goal. \MAPF solution paths are shown in pink.}
    \label{fig:ngr}
\end{figure}

We call our algorithm \MfM: Multi-Agent Pathfinding for Manipulation Among Movable Objects.
\MfM is given access to a physics-based simulator (PyBullet~\cite{coumans2019}) to ensure that no interaction constraints defined in the \MAMO problem are violated.
We note that a \MAMO problem $\PP$ to retrieve the OoI with $\OM \neq \emptyset$ is solvable \emph{iff} the simpler problem $\hat{\PP}$ without any movable objects i.e., $\OM = \emptyset$ can be solved.
We denote a solution trajectory to $\hat{\PP}$ as $\hat{\pi}_\R$.
Let $\mathcal{V}\left(\hat{\pi}_\R\right)$ denote the volume occupied by the robot arm in the workspace during execution of $\hat{\pi}_\R$.
$\mathcal{V}\left(\hat{\pi}_\R\right)$ specifies a ``negative goal region'' (NGR)~\cite{DogarS12} for the movable objects.
A NGR is a sufficient volume of the 3D workspace which, if there are no objects inside it, allows the robot arm to retrieve the OoI without other contacts.
If all movable objects can be rearranged such that they are outside $\mathcal{V}\left(\hat{\pi}_\R\right)$, the robot can execute $\hat{\pi}_\R$ to retrieve the OoI.
Fig.~\ref{fig:ngr} shows a NGR $\mathcal{V}\left(\hat{\pi}_\R\right)$ for the problem from Fig.~\ref{fig:100011}.

Algorithm~\ref{alg:mfm} contains the pseudocode for \MfM. At a high-level, \MfM first computes $\hat{\pi}_\R$ (Line~\ref{line:simple_mamo}) and the NGR $\mathcal{V}\left(\hat{\pi}_\R\right)$ (Line~\ref{line:ngr}).
It then iterates over two steps:
\begin{enumerate}
    \item\label{bullet:call_mapf} Section~\ref{sec:key_idea}: Compute a solution to the abstract Multi-agent Pathfinding (\MAPF) problem where each movable object is treated as an agent that needs to escape the NGR without colliding with other agents using Conflict-Based Search (\CBS)~\cite{SharonSFS15}, a complete and optimal \MAPF algorithm.
    \item Section~\ref{sec:push_actions}: Pick a movable object to be rearranged according to the \MAPF plan computed in~\ref{bullet:call_mapf} and find a valid non-prehensile push for it by forward simulating potential pushes using a physics-based simulator.
\end{enumerate}

Algorithm~\ref{alg:mfm} uses \texttt{replan} to ensure \CBS is only called to solve new \MAPF problems.
After the first \CBS call, \texttt{replan} triggers subsequent \CBS calls once a valid push has been found i.e., at least one object has been moved.
This leads to a different \MAPF problem with new object poses.
Until a valid push is found, we sample and simulate pushes for all objects that move in the \MAPF solution.





The \textsc{PlanRetrieval} function takes as input a set of objects to be considered as immovable obstacles for the robot and runs Multi-Heuristic A$^\ast$~\cite{AineSNHL16} to find an arm trajectory in $\X_\R$ to retrieve the OoI.

\CBS is called in Line~\ref{line:plan_mapf} with the latest known movable object poses in $SE(3)$ to obtain a set of paths that ensure they all satisfy the NGR $\mathcal{V}\left(\hat{\pi}_\R\right)$.
This searches over all possible rearrangements of the scene from the current state, without ever querying a physics simulator, by assuming that movable objects are artificially actuated agents (Section~\ref{sec:key_idea}).

We then loop over all objects that need to be rearranged (from Line~\ref{line:plan_rearrange}) and try and find a valid push for them (Section~\ref{sec:push_actions}).
If a valid push is found (Line~\ref{line:push_found}), it is added to the final sequence of arm trajectories to be executed $\Psi$, and the pose of that object is updated for future iterations.

\MfM terminates either when the allocated planning budget expires, or we successfully find a trajectory to retrieve the OoI in the presence of all objects ($\OI \cup \OM$) as obstacles in Line~\ref{line:finalise_plan}.
Although this trajectory $\pi_\R$ may be different from $\hat{\pi}_\R$ (Line~\ref{line:simple_mamo}), it will still retrieve the OoI successfully since it is guaranteed to not make contact with any object (immovable or movable).
The sequence of trajectories $\Psi$ can then be executed in order to rearrange the movable objects (if required) and finally ending in successful OoI retrieval.

\begin{algorithm}[t]
\begin{small}
\caption{Multi-Agent Pathfinding for Manipulation Among Movable Objects}\label{alg:mfm}
\begin{algorithmic}[1]

\Procedure{\MfM}{$\OM^{\text{init}}, \OI$}
    \State $\OM \gets \OM^{\text{init}}$ \Comment{Rearranged object positions}
    \State $\Psi \gets \emptyset$ \Comment{Sequence of arm trajectories}
    \State $\hat{\pi}_{\R} \gets \textsc{PlanRetrieval}(\OI)$ \label{line:simple_mamo} \Comment{OoI retrieval trajectory}
    \State Compute $\mathcal{V}(\hat{\pi}_{\R})$ \label{line:ngr}
    \State \texttt{replan} $\gets$ \texttt{true}, \texttt{done} $\gets$ \texttt{false}
    \While{\texttt{time remains}}
        \If{\texttt{replan}} \label{line:push_found}
            \State $\pi_{\R} \gets \textsc{PlanRetrieval}(\OI \cup \OM)$ \label{line:finalise_plan}
            \If{$\pi_{\R}$ exists}
                \State $\Psi \gets \Psi \cup \{\pi_{\R}\}$, \texttt{done} $\gets$ \texttt{true}
                \State \textbf{break} \label{line:finalised}
            \EndIf
            \State $\{\pi_{o_m}\}_{o_m \in \OM} \gets \CBS( \OM, \OI, \mathcal{V}(\hat{\pi_{\R}}))$ \label{line:plan_mapf}
            \State \texttt{replan} $\gets$ \texttt{false}
        \EndIf
        \For{$o_m \in \OM$} \label{line:plan_rearrange}
            \If{$\pi_{o_m} = \emptyset$} \label{line:dont_move}
                \State \textbf{continue}
            \EndIf
            \State $\psi \gets \textsc{PlanPush}(o_m, \pi_{o_m}, \OM, \OI)$ \label{line:plan_push}
            \State $(\texttt{valid}, o_m^\prime) \gets \textsc{SimulatePush}(\psi)$ \label{line:sim_push}
            \If{\texttt{valid}} \label{line:push_found}
                \State $\Psi \gets \Psi \cup \{\psi\}$, \texttt{replan} $\gets$ \texttt{true}
                \State $\textsc{UpdatePose}(\OM, o_m^\prime)$
                \State \textbf{break}
            \EndIf
        \EndFor
    \EndWhile
    \If{$\neg\texttt{done}$}
        \Return $\emptyset$
    \EndIf
    \Return $\Psi$
\EndProcedure

\end{algorithmic}
\end{small}
\end{algorithm}


\subsection{\MAPF Abstraction for Manipulation}\label{sec:key_idea}

A fundamental challenge to solving \MAMO problems requires determining \emph{which} objects need to be rearranged and \emph{where} they should be moved.
The key idea in this paper uses an existing \MAPF solver to search over potential rearrangements of the scene which lead to successful OoI retrieval.
Importantly, the \MAPF solver does not require access to a physics simulator for this purpose -- it only relies on 3D collision checking.
Our \MAPF abstraction includes all movable objects $o_m \in \OM$ as agents.
We check for collisions between agents in space and time in their full $SE(3)$ configuration space.
All agents have a discrete action space corresponding to a four-connected grid on the $(x, y)-$plane of the shelf.
We assume each action takes unit time and either the agent remains in place, or the $x-$ or $y-$coordinate of the agent pose changes by $\SI{1}{\centi\meter}$.

Agent start configurations are determined by their latest pose in $SE(3)$ prior to the \MAPF call (Algorithm~\ref{alg:mfm}, Line~\ref{line:plan_mapf}).
Each agent $o_m$ in the \MAPF problem has a set of possible goals that include all states where the agent satisfies the NGR by being ``outside'' it.

We call \CBS to obtain a solution, shown in Fig.~\ref{fig:ngr}, to this \MAPF abstraction.
The solution is a set of paths for movable objects $\{\pi_{o_m}\}_{o_m \in \OM}$ whose final states $\pi_{o_m}^{end}$ satisfy the NGR, and suggests a rearrangement strategy in terms of \emph{which} objects to move and \emph{where}.
If we can rearrange all $o_m \in \OM$ to their respective $\pi_{o_m}^{end}$ poses, we know that the trajectory $\hat{\pi}_\R$ will successfully retrieve the OoI, thereby solving the \MAMO problem.

\subsection{Generating Non-Prehensile Push Actions}\label{sec:push_actions}
\begin{figure}[t]
    \centering
    \includegraphics[width=0.95\columnwidth]{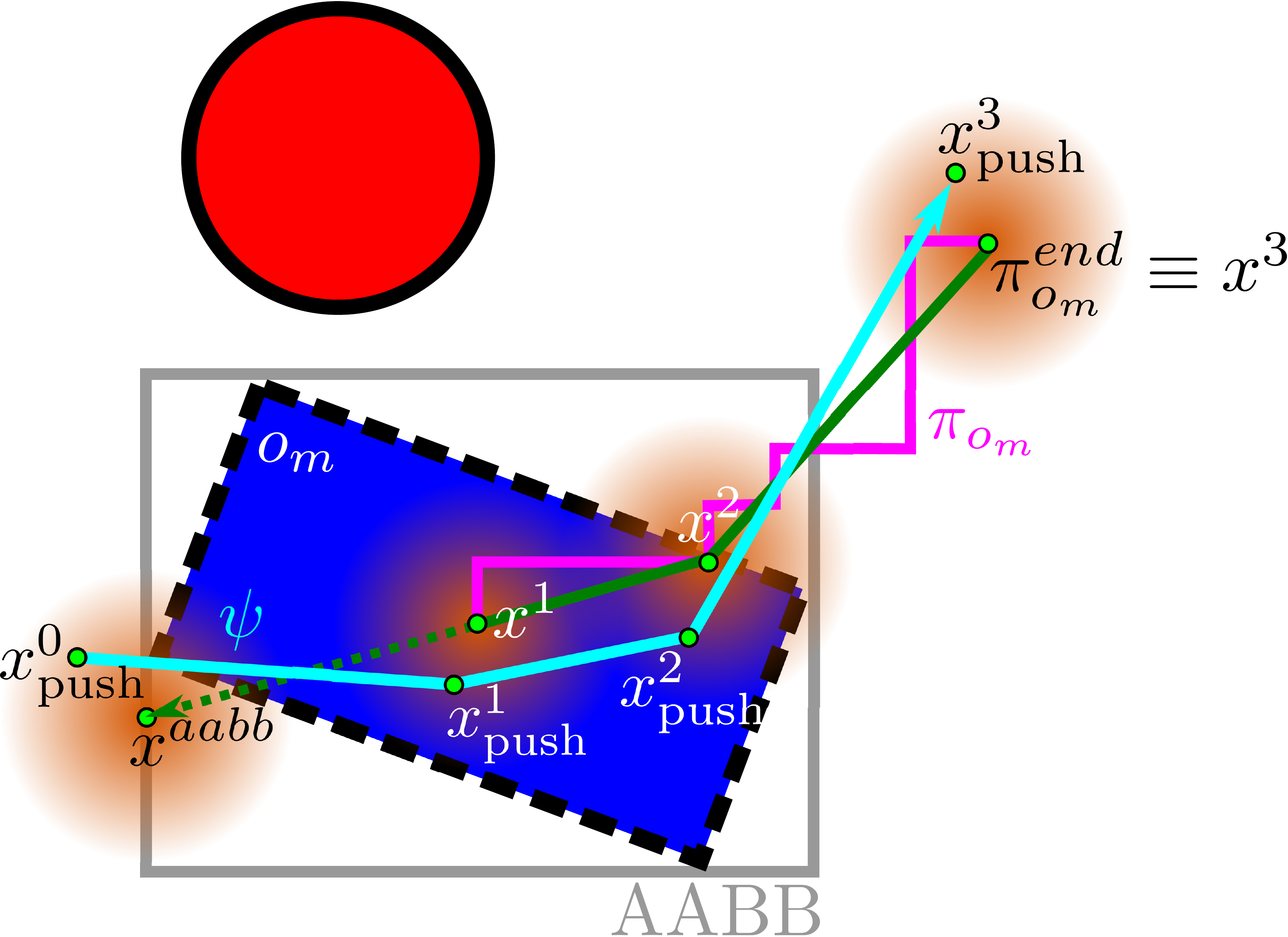}
    \caption{2D illustration of our push planner. Given a movable object $o_m$ (blue) and its \MAPF solution path $\pi_{o_m}$ (pink), we shortcut $\pi_{o_m}$ while accounting for immovable obstacles $\OI$ (red) to get the green path of straight line segments. After computing $x^{aabb}$ by intersecting the $\protect\overrightarrow{(x^2, x^1)}$ ray with the axis-aligned bounding box for $o_m$, the push action (cyan) is computed via inverse kinematics between sampled points $x_{\text{push}}^{i} \sim \mathcal{N}(x^i, \sigma I), \,i = \{0,\ldots,n\}, \, x^0 \vcentcolon= x^{aabb}$.}
    \label{fig:push_algo}
\end{figure}

Given a path $\pi_{o_m}$ for $o_m \in \OM$ from the \MAPF solution, \textsc{PlanPush} (Algorithm~\ref{alg:mfm}, Line~\ref{line:plan_push}) determines \emph{how} an object may be rearranged (Fig.~\ref{fig:push_algo}).
We would like to move the object to $\pi_{o_m}^{end}$, which is known to satisfy the NGR.
To compute a push trajectory, we first shortcut $\pi_{o_m}$ (taking into account collisions with immovable obstacles $\OI$) into a series of straight line segments defined by points $\{x^1 = \pi_{o_m}^{start}, \ldots, x^n = \pi_{o_m}^{end}\}$.
We also compute the point of intersection $x^{aabb}$ of the ray from $x^1$ along the direction $\overrightarrow{(x^2, x^1)}$ with the axis-aligned bounding box of $o_m$.

\textsc{PlanPush} computes a collision-free path between successive pushes by planning in $\X_\R$ with all objects $\OI \cup \OM$ as obstacles to a point $x_{\text{push}}^{0}$ sampled around $x^{aabb}$\footnote{We sample $(x, y)$ coordinates for $x_{\text{push}}^{0}$ from $\mathcal{N}(x^{aabb}, \sigma I), \,\sigma = \SI{2.5}{\centi\meter}$. The $z-$coordinate is fixed at $\SI{3}{\centi\meter}$ above the shelf for the entire push action.}.
If this path is found, \textsc{PlanPush} similarly samples points $x_{\text{push}}^{i}$ around each $x^i$ in the shortcut path.
It runs inverse kinematics (IK) in sequence for each segment of the push action between points $\left(x_{\text{push}}^{i-1}, x_{\text{push}}^{i}\right), i = \{1, \ldots, n\}$.
If all IK calls succeed, we return the full push trajectory by concatenating $\pi_0$ with all push action segments.

This push action, informed by the \MAPF solution about \emph{which} object to move \emph{where}, is forward simulated with a physics model to verify whether it satisfies all interaction constraints for all objects.
If so, it is queued into the sequence of rearrangements that will be executed as part of the \MAMO solution returned by \MfM (Algorithm~\ref{alg:mfm}, Line~\ref{line:push_found}).




\section{Experimental Results}\label{sec:exps}


\subsection{Simulation Experiments}\label{sec:sim_exps}

\begin{figure}[t]
    \centering
    \includegraphics[width=0.95\columnwidth]{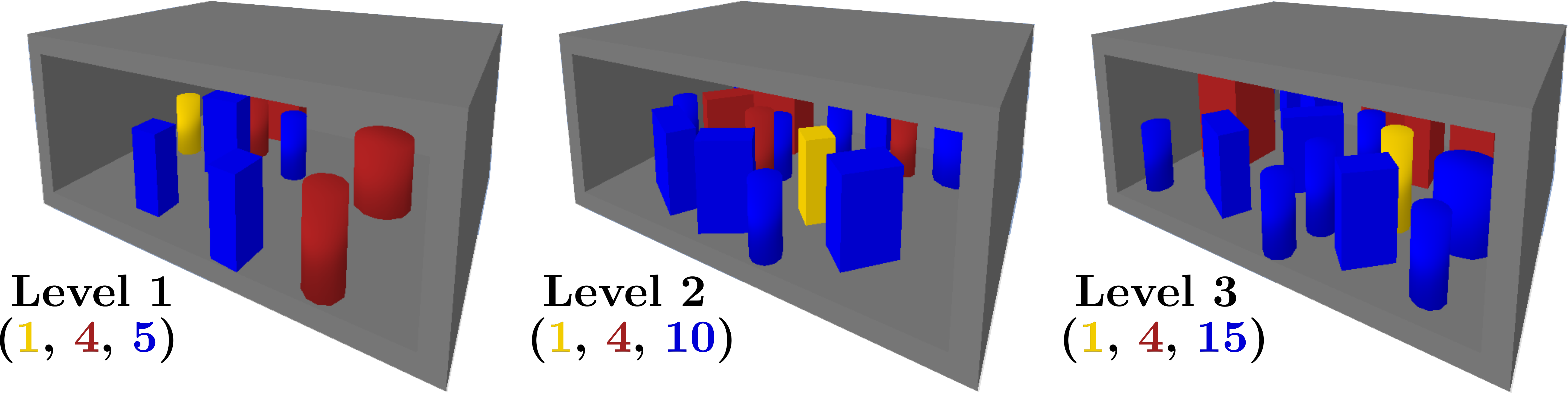}
    \caption{\MAMO problems of differing complexity. From \emph{left} to \emph{right}, Levels 1, 2, and 3 have 5, 10, and 15 movable objects respectively. Each Level has 1 OoI and 4 immovable obstacles.}
    \label{fig:levels}
\end{figure}

We run our simulation experiments in \MAMO workspaces of three difficulty levels shown in Fig.~\ref{fig:levels}.
Each workspace has one OoI (yellow), four immovable obstacles (red), and different numbers of movable objects (blue).
Objects are cylinders and cuboids with random sizes, initial poses, masses, and coefficients of friction.
We assume perfect knowledge of the initial workspace state and all object parameters.
We set a planning timeout of $\SI{120}{\second}$ for 100 randomly generated \MAMO problems at each level.
Our analysis includes two versions of our algorithm -- \MfM refers to Algorithm~\ref{alg:mfm}, and $\widehat{\MfM}$ refers to a version which only calls \CBS once (after Line~\ref{line:ngr}) and does not iterate between calling \CBS and finding a valid push in simulation.




\textbf{Baselines: }
We compare the performance of \MfM against three types of baselines for solving \MAMO problems with non-prehensile interactions.
The first are standard implementations of sampling-based algorithms \KPIECE~\cite{KPIECE} and RRT~\cite{RRT} from OMPL~\cite{OMPL} that search the entire \MAMO state space $\X$ by randomly sampling robot motions.

The second baseline, Selective Simulation~\cite{SelSim} (\SelSim), is a search-based algorithm that interleaves a `planning' phase and a `tracking' phase.
The former queries the physics-based simulator for interactions with a set of `relevant' movable objects identified so far.
The latter executes the solution found by the planning phase in the presence of all objects in simulation and, if any interaction constraints are violated, it adds the `relevant' object to the set.
It only uses the simple motion primitives described in Section~\ref{sec:prelims}.

Our final baseline is the work from Dogar et al.~\cite{DogarS12} (\Dogar) which introduced the idea of a negative goal region (NGR) we use in \MfM.
\Dogar recursively searches for a solution backwards in time, similar to~\cite{StilmanMAMO}.
It first finds an OoI retrieval trajectory ignoring all movable objects.
The NGR induced by this trajectory helps identify a set of objects to be rearranged, and the OoI is added as an obstacle.
If an object is successfully rearranged, the NGR and set of objects still to be rearranged are updated with the trajectory found, and the rearranged object is added as an obstacle at its initial pose.
This process continues until no further objects need to be rearranged. 
Our implementation of \Dogar finds the same OoI retrieval trajectory as \MfM, and uses the same push actions (Section~\ref{sec:push_actions}) to try and rearrange objects.
Notably, \Dogar only has information about \emph{which} objects to move but not \emph{where} to move them.
Our implementation finds the closest cell outside the latest NGR for an object and samples points around this location to try to move the object towards.



\begin{table*}
\centering
\caption{Simulation Study for \MAMO Planning in Cluttered Scenes - success rates and \emph{min/median/max} planning and simulation times}
\label{tab:sim_main}
\begingroup
\begin{tabular}{@{}cccccccc@{}}
\toprule
\multirow{2}{*}{\textbf{Metrics}} & \multirow{2}{*}{\textbf{Level}} & \multicolumn{6}{c}{\textbf{Planning Algorithms}} \\
\cmidrule{3-8}
& & \textbf{\MfM} & $\widehat{\MfM}$ & \Dogar~\cite{DogarS12} & \SelSim~\cite{SelSim} & \KPIECE~\cite{KPIECE} & \RRT~\cite{RRT}  \\ \midrule
\multirow{3}{*}{\shortstack{Success\\ Rate $(\%)$}} & 1 & 92 & 79 & 40 & 33 & 48 & 55 \\
 & 2 & 73 & 54 & 20 & 21 & 33 & 40 \\
 & 3 & 62 & 36 & 6 & 16 & 17 & 26 \\ \midrule
\multirow{3}{*}{\shortstack{Total\\ Planning\\ Time $(\SI{}{\second})$}} & 1 & 1.0 / 2.6 / 102.5 & 1.0 / 2.4 / 103.8 & 0.1 / 0.9 / 115.3 & 0.004 / 0.02 / 0.03 & 7.4 / 23.4 / 117.8 & 7.1 / 15.8 / 101.4 \\
 & 2 & 1.2 / 6.6 / 115.4 & 1.3 / 2.6 / 100.3 & 0.3 / 0.5 / 113.5 & 0.002 / 0.008 / 0.2 & 9.3 / 28.2 / 112.0 & 8.6 / 27.6 / 104.9 \\
 & 3 & 1.3 / 7.2 / 116.1 & 1.6 / 2.4 / 72.6 & 0.2 / 0.4 / 55.0 & 0.004 / 0.01 / 0.03 & 10.6 / 32.0 / 98.5 & 10.3 / 26.7 / 113.4 \\ \midrule
\multirow{3}{*}{\shortstack{Simulation\\ Time $(\SI{}{\second})$}} & 1 & 0 / 0 / 58.6 & 0 / 0 / 20.1 & 0 / 0 / 42.0 & 27.3 / 35.0 / 43.6 & 0 / 10.6 / 99.0 & 0 / 4.4 / 87.2 \\
 & 2 & 0 / 0.4 / 75.9 & 0 / 0 / 37.0 & 0 / 0 / 20.9 & 36.7 / 44.1 / 58.3 & 0 / 16.1 / 95.4 & 0 / 16.7 / 83.7 \\
 & 3 & 0 / 0.4 / 55.1 & 0 / 0 / 24.3 & 0 / 0 / 20.0 & 47.3 / 55.7 / 76.0 & 0 / 18.3 / 79.3 & 0 / 15.3 / 101.2 \\ \bottomrule
\end{tabular}
\endgroup
\end{table*}

\begin{figure*}[t]
    \centering
    \includegraphics[width=\textwidth]{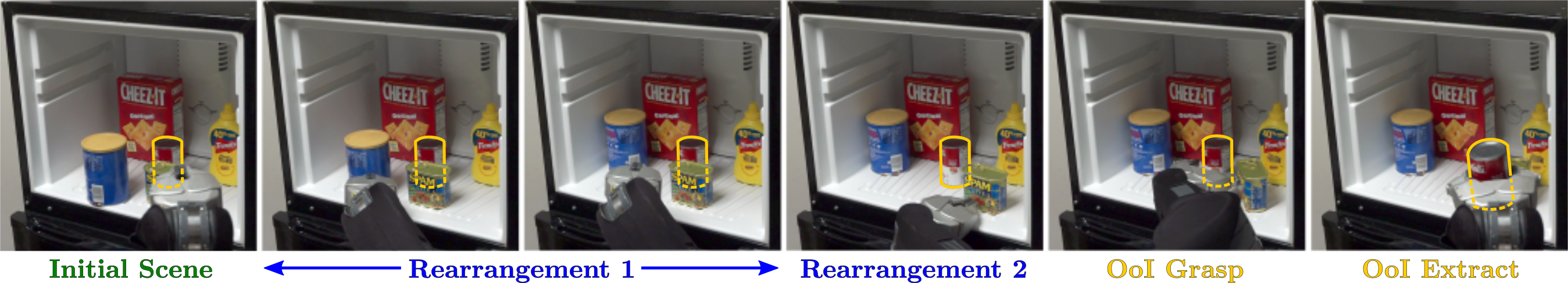}
    \caption{A \MAMO solution generated by \MfM. The tomato soup can (yellow outline) is the OoI, all other objects are movable.}
    \label{fig:pr2_soln}
\end{figure*}


\textbf{Results: }
Table~\ref{tab:sim_main} shows the result of our experiments where we present the \emph{min}$/$\emph{median}$/$\emph{max} values for total planning time and simulation time of successful runs only.
Experiments were run on a $\SI{4}{\giga\hertz}$ Intel i7-4790K CPU with $\SI{28}{\giga\byte}$ $\SI{1600}{\mega\hertz}$ DDR3 RAM.

Both versions of \MfM solve the most problems across all difficulty levels.
For Levels 1, 2, and 3, the \MfM solution successfully executed $0.8$, $1.9$, and $3.1$ push actions on average.
The difference in performance between \MfM and $\widehat{\MfM}$ highlights the benefit of the iterative nature of \MfM.
Since \MAPF paths are usually not precisely replicated in simulation via pushes, querying the solver repeatedly with an updated workspace configuration leads to more informed future paths for objects, instead of trying to forcibly push them to the first goal configuration suggested by \MAPF.

All baseline algorithms from Table~\ref{tab:sim_main} suffer due to poor exploration over the space of rearrangements. 
Our approach benefits from the \MAPF abstraction to produce guidance on \emph{where} to move each object to free up the NGR.
The stochastic sampling of push actions used by our push planner leads to complex, multi-body non-prehensile interactions that satisfy interaction constraints in the final solution.
In contrast \Dogar naively samples pushes to be simulated, and necessarily tries to ensure there is no overlap between the NGR and movable objects, even if a slightly different collision-free path can be found to retrieve the OoI (Algorithm~\ref{alg:mfm}, Line~\ref{line:finalise_plan}).
This strategy suffers when sampled points are near immovable obstacles, and limits the possible rearrangements considered since movable objects that are rearranged successfully are treated as immovable obstacles.
\Dogar also never executes a potential trajectory until there is no overlap between the NGR and movable objects, unlike \SelSim which simulates all trajectories found during planning.
In fact, all \SelSim successes in Table~\ref{tab:sim_main} correspond to scenes where the very first planned trajectory succeeds in OoI retrieval in simulation.
This is only true when there is minimal overlap between the NGR and movable objects.
When any movable object needs to be rearranged, \SelSim suffers from its poor action space -- the simple motion primitives are ineffective at causing meaningful robot-object interactions in the workspace. 
\KPIECE and \RRT benefit significantly from goal biasing in simpler scenes where either little to no robot-object interactions are required or the objects that need to be moved have nice physical properties (large supporting footprint, low center-of-mass, low coefficient of friction).

\subsection{Real-World Performance on the PR2}\label{sec:pr2_exps}

We ran \MfM on a PR2 robot where we used a refrigerator compartment as our \MAMO workspace (Fig.~\ref{fig:pr2_soln}).
We placed five objects from the YCB Object Dataset~\cite{YCB} in the refrigerator.
Four of these were movable and the tomato soup can was the object-of-interest.
Objects were localised using a search-based algorithm~\cite{Agarwal2020PERCH2} run on a NVidia Titan X GPU.
We gave \MfM a total planning timeout of $\SI{120}{\second}$.


Out of 16 perturbations of the initial scene from Fig.~\ref{fig:pr2_soln}, 12 runs successfully retrieved the OoI.
Across the successful runs the planner took $56.41 \pm 27.29 \,\SI{}{\second}$ to compute a plan of which $49.26 \pm 24.21 \,\SI{}{\second}$ was spent simulating pushes.
Failures were due to interaction constraints being violated during execution by the PR2.
Since \MfM returns a solution that does not violate constraints in simulation, failures are due to modelling errors between the simulator and the real-world.
Specifically, accurately computing coefficients of friction is difficult and can lead to differing contact mechanics in simulation than the real-world.
Fig.~\ref{fig:pr2_soln} shows the solution to a \MAMO problem being executed by the PR2.
It moves the coffee can out of the way, pushes the potted meat can slightly aside, and finally the OoI (tomato soup can) is extracted while also nudging the potted meat can.

\section{Conclusion and Discussion}\label{sec:discussion}

This paper presents \MfM: Multi-Agent Pathfinding for Manipulation Among Movable Objects, an algorithm to plan for manipulation in heavy clutter that considers complex interactions such as rearranging multiple objects simultaneously, and tilting, leaning and sliding objects.
\MfM uses a \MAPF abstraction to the \MAMO problem to find suitable rearrangements, and a non-prehensile push planner to realise these rearrangements by utilising complex multi-body interactions.
It dramatically outperforms alternative approaches that do not reason about such interactions efficiently.

\MfM greedily commits valid pushes found to its sequence of rearrangement trajectories.
This greedy behaviour makes \MfM incomplete, given that it has no ability to backtrack from this decision.
In the future we hope to address this incompleteness of \MfM by developing an algorithm that considers (i) all feasible pushes for an object that needs to be rearranged to a specific location, (ii) all orderings of all feasible push actions to realise a particular rearrangement for a set of objects, and (iii) all possible rearrangements for a set of objects.
Additionally, the \MAPF solver used in \MfM should be modified to use a cost function which has information about robot kinematics and pushing dynamics so as to compute and thus simulate better push actions.


\balance
\bibliographystyle{IEEEtran}
\bibliography{references}

\end{document}